\documentclass[conference]{IEEEtran}
\IEEEoverridecommandlockouts
\usepackage{cite}
\usepackage{amsmath,amssymb,amsfonts}
\usepackage{algorithmic}
\usepackage{graphicx}
\usepackage{textcomp}
 \usepackage{multirow}
 \usepackage{adjustbox}
 \usepackage{booktabs} 
\usepackage{xcolor}
\def\BibTeX{{\rm B\kern-.05em{\sc i\kern-.025em b}\kern-.08em
    T\kern-.1667em\lower.7ex\hbox{E}\kern-.125emX}}
\begin{document}

\title{Mitigating Hallucination on Hallucination in RAG via Ensemble Voting    \\
}

\author{\IEEEauthorblockN{Zequn Xie}
\IEEEauthorblockA{\textit{Polytechnic Institute} \\
\textit{Zhejiang University}\\
 Hangzhou, China }

\and
\IEEEauthorblockN{ Zhengyang Sun}
\IEEEauthorblockA{\textit{ School of Computer Science and Technology} \\
\textit{Xinjiang University}\\
Urumqi, China \\
107552404542@stu.xju.edu.cn}

}

\maketitle

\begin{abstract}
Retrieval-Augmented Generation (RAG) aims to reduce hallucinations in Large Language Models (LLMs) by integrating external knowledge. However, RAG introduces a critical challenge: hallucination on hallucination,'' where flawed retrieval results mislead the generation model, leading to compounded hallucinations. To address this issue, we propose \textbf{VOTE-RAG}, a novel, training-free framework with a two-stage structure and efficient, parallelizable voting mechanisms. VOTE-RAG includes: (1) \textit{Retrieval Voting}, where multiple agents generate diverse queries in parallel and aggregate all retrieved documents; (2) \textit{Response Voting}, where multiple agents independently generate answers based on the aggregated documents, with the final output determined by majority vote. We conduct comparative experiments on six benchmark datasets. Our results show that VOTE-RAG achieves performance comparable to or surpassing more complex frameworks. Additionally, VOTE-RAG features a simpler architecture, is fully parallelizable, and avoids the problem drift'' risk. Our work demonstrates that simple, reliable ensemble voting is a superior and more efficient method for mitigating RAG hallucinations.

\end{abstract}
\section{Introduction}

Large Language Models (LLMs) have demonstrated remarkable natural language understanding and reasoning capabilities \cite{achiam2023gpt, touvron2023llama}. \textbf{To better understand the internal mechanisms of these models and the origins of such errors, recent surveys have emphasized the importance of explainability in tracing the decision-making process of LLMs \cite{palikhe2025towards}.} However, their reliance on parametric knowledge introduces a critical challenge: hallucination, where the generated content deviates from factual correctness \cite{ji2023survey, huang2024survey}. This issue severely limits their reliability, particularly in knowledge-intensive tasks where factual accuracy is paramount. \textbf{This phenomenon is akin to the challenges faced in multi-modal domains, where foundational models must be carefully designed to ensure cross-modal alignment and factual grounding \cite{yu2026dinov3}.} To mitigate hallucinations, Retrieval-Augmented Generation (RAG) has been proposed as a framework that integrates external knowledge retrieval to enhance LLM outputs \cite{gao2023retrieval}. By conditioning responses on retrieved documents, RAG reduces reliance on the model’s parametric knowledge, thereby aiming to improve factual correctness.

RAG introduces a new challenge: biased or erroneous retrieval results can mislead generation and amplify the LLM’s inherent hallucinations, producing what is called \textit{Hallucination on Hallucination} \cite{drag}. Addressing this challenge requires optimizing the entire RAG pipeline — both retrieval and generation. In retrieval, incomplete or biased results mislead the generation stage; even when retrieval strategies are improved, the generator remains vulnerable to retrieval noise and misinformation. \textbf{The principles of reasoning in computer vision provide a structured foundation for understanding how models parse such complex relations and noise in various modalities \cite{sarkar2025reasoning}.} 

Recent research has explored various optimization techniques to enhance the precision of the retrieval process. For instance, chat-driven interactions have been leveraged to improve text generation and interaction for person retrieval \cite{xie2025chat}, and context-aware representations with query enhancement have shown significant promise in reducing search ambiguity \cite{xie2026conquer}. \textbf{Furthermore, just as RAG requires noise reduction in text, specialized image-to-image translation and visual imperfection removal techniques have been developed to ensure that generated or retrieved visual content remains factually and aesthetically unblemished \cite{yu2025yielding, yu2025qrs}.} Advancements in human vision-driven video representation \cite{xie2026hvd} and hierarchical visual perception \cite{xie2026delving} demonstrate that incorporating multi-level semantic alignment can further stabilize retrieval in complex domains. Despite these advancements, existing RAG mitigations—such as iterative retrieval \cite{trivedi2023interleaving, shao2023enhancing, feng2024retrieval} and reflection during generation \cite{asai2023self}—typically rely on single-agent self-optimization and therefore remain susceptible to inherent biases. \textbf{This challenge of managing specific model biases is also reflected in the context of selective forgetting, where benchmarking a model's capability to manage specific information is crucial for maintaining overall reliability \cite{yu2025forgetme}.}

To address the two-stage challenge of \textit{Hallucination on Hallucination}, a robust and efficient mechanism is required to improve both retrieval quality and generation fidelity. Instead of relying on a single, complex self-correction loop, we turn to the principles of ensemble methods. The ``wisdom of the crowd,'' realized through aggregation, is a powerful paradigm for enhancing robustness and mitigating errors from individual components. In multi-agent systems, this principle has proven to be highly effective. Recent research \cite{choi2025debate} provides strong empirical evidence that simple aggregation mechanisms, such as majority voting, are primary drivers of performance gains and decision accuracy in multi-agent LLM systems. \textbf{Such robust retrieval and estimation strategies are particularly effective in specialized domains, such as remote sensing, where terrain-aware and spatiotemporal alignment are necessary to recover accurate information from noisy data sources \cite{yu2026spatiotemporal}.}

Motivated by the power and simplicity of ensemble voting, we propose \textbf{VOTE-RAG}, a training‑free framework that applies this principle to both stages of the RAG pipeline. \textbf{Our approach to modular aggregation mirrors recent trends in training-free multilingual text editing, which utilizes layered disentanglement to achieve high-fidelity results without extensive retraining \cite{yu2025cotextor}.} VOTE-RAG operates in two phases to target both sources of error. In the retrieval phase, \textit{Retrieval Voting} has multiple agents generate diverse queries in parallel and aggregates all retrieved documents into a shared pool, broadening knowledge coverage and reducing bias. In the generation phase, \textit{Response Voting} uses $N$ independent agents to produce answers from the same aggregated document pool, and the final answer is selected by simple majority vote. 

Overall, our main contributions are summarized as follows:
\begin{itemize}
    \item We introduce \textbf{VOTE-RAG}, a novel training-free framework that integrates ``Ensemble Voting'' mechanisms into both the retrieval and generation phases of RAG.
    \item We evaluate VOTE-RAG on multiple tasks, demonstrating its effectiveness in improving retrieval reliability, enhancing reasoning robustness, and substantially reducing RAG-induced hallucinations.
\end{itemize}
\section{Related Work}

\subsection{Limitations of Standard RAG}
LLMs have demonstrated remarkable success across a wide range of tasks \cite{achiam2023gpt, touvron2023llama}. 
However, they are prone to hallucinations, particularly when handling queries beyond their training data \cite{ji2023survey, huang2024survey}.
To mitigate this issue, Retrieval-Augmented Generation (RAG) enhances LLMs by incorporating external knowledge retrieval into the response generation process \cite{gao2023retrieval}.
The most straightforward approach involves using the initial input as a query to retrieve information from an external corpus \cite{guu2020retrieval, lewis2020retrieval, izacard-grave-2021-leveraging, izacard2022few, shi-etal-2024-replug}. 
However, this standard RAG pipeline, which relies on a single query and a single generation pass, has key weaknesses. As noted in \cite{drag}, this process is vulnerable to ``Hallucination on Hallucination'': \textbf{Retrieval Failure:} A single-pass retrieval based solely on the initial input may fail to retrieve essential knowledge or may fetch biased, irrelevant information, leading to a flawed context. \textbf{Generation Failure:} Even if the correct information is retrieved, it is often mixed with noise. A single generation pass may be influenced by this noise, leading to an incorrect answer. \textbf{Such reliability issues are further compounded by the lack of transparency in how LLMs process retrieved evidence, necessitating a deeper exploration of explainable AI to trace and correct these factual deviations \cite{palikhe2025towards}.}
While methods like iterative retrieval \cite{trivedi2023interleaving, shao2023enhancing} and self-reflection \cite{asai2023self} exist, these approaches still rely on a single-agent reasoning process, which can be susceptible to its own inherent biases and failure modes \cite{wei2008fusing}. \textbf{This challenge of managing specific agent biases is also observed in the context of selective forgetting, where maintaining a model's reliability requires a precise balance between retained and discarded information \cite{yu2025forgetme}.}

\subsection{Multi-Agent Ensembles and Voting} 
To overcome the limitations of single-agent approaches, leveraging multi-agent systems is a promising direction. Aggregating outputs from multiple independent agents mitigates individual biases and substantially improves reasoning robustness \cite{becker2024multi}. Recent work has focused on finding the most effective and efficient aggregation mechanism; a critical study by Choi et al. \cite{choi2025debate} provides strong empirical evidence that the primary driver of gains is \textbf{majority voting}. They show that aggregating independent agent responses is a simple, robust, and reliable method that often outperforms more complex interaction strategies. \textbf{This trend toward efficient, training-free aggregation is similarly reflected in multilingual text editing, where modular disentanglement and fusion provide high-fidelity results without the need for iterative fine-tuning \cite{yu2025cotextor}.} This insight---that the strength of multi-agent systems lies in straightforward, parallel aggregation---underpins our approach. Rather than pursuing complex iterative self-correction, \textbf{VOTE-RAG} leverages ensemble voting to directly address the two-stage ``Hallucination on Hallucination'' problem in RAG.

\subsection{Robustness and Disambiguation in Complex Retrieval}
While text-based RAG deals with inherent ambiguity and noise in retrieved contexts, similar challenges have been extensively studied and addressed in complex multi-modal retrieval paradigms, such as Composed Image Retrieval (CIR) and Composed Video Retrieval (CVR). \textbf{A comprehensive taxonomy of reasoning in computer vision highlights the importance of multi-step methodologies in resolving such cross-modal ambiguities \cite{sarkar2025reasoning}.} In these domains, queries consist of hybrid inputs (e.g., a reference image combined with a text modifier), making the retrieval process highly susceptible to noise, semantic misalignment, and focus shifts. Recent advancements in CIR and CVR provide valuable insights into robust retrieval mechanisms. For instance, approaches have been developed to explicitly parse fine-grained modification semantics \cite{FineCIR} and bind entity-modification relations \cite{ENCODER} to ensure highly precise retrieval targets. \textbf{Such precision is equally critical in quantitative remote sensing, where foundation models must be robustly aligned to recover accurate information from terrain-aware and spatiotemporal data \cite{yu2026dinov3, yu2026spatiotemporal}.} 

To mitigate the impact of inherent noise and ambiguity, researchers have proposed hierarchical uncertainty-aware disambiguation networks \cite{HUD} and invariance-aware noise mitigation strategies \cite{INTENT}. \textbf{In cases where the retrieved or generated visual context is noisy, advanced visual imperfection removal techniques and style-transfer-based translations can be utilized to maintain aesthetic and factual consistency \cite{yu2025yielding, yu2025qrs}.} Furthermore, addressing spatial and semantic focus shifts during retrieval has been explored through segmentation-based revision \cite{OFFSET}, while other methods enhance overall retrieval accuracy by carefully separating shared and differential semantics \cite{REFINE}. The core principles of these techniques---specifically, rigorous semantic parsing, uncertainty disambiguation, and noise reduction---are highly relevant to addressing the ``Retrieval Failure'' in standard RAG, providing a foundation for more resilient context-fetching mechanisms prior to generation.

\begin{figure*}
    \centering
     \includegraphics[width=1.0\linewidth]{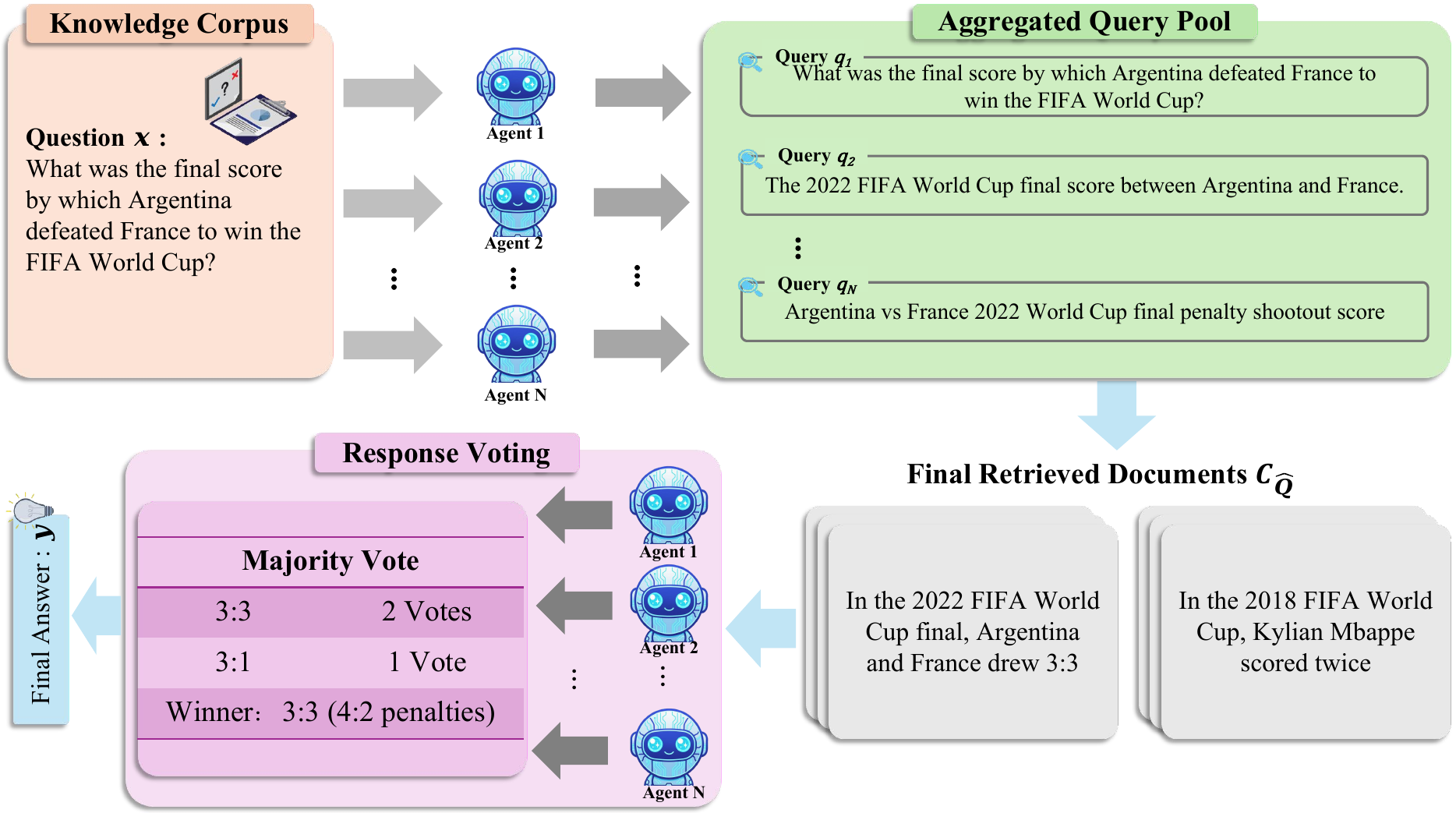} % 假设您有了一张 VOTE-RAG 的新图
    \caption{An overview of our VOTE-RAG framework. It leverages parallel ensemble voting to enhance retrieval breadth and generation robustness against "Hallucination on Hallucination."}
    \label{fig:framework}
\end{figure*}
\section{THE VOTE-RAG FRAMEWORK}
\label{sec:framework}

In this section, we introduce VOTE-RAG, a general framework designed to enhance Retrieval-Augmented Generation (RAG) through a structured ensemble voting mechanism. We first provide an overview of VOTE-RAG in Section~\ref{sec:method_overview}, followed by a comprehensive discussion of its components. As illustrated in Fig. 1, VOTE-RAG consists of a two-stage voting framework that comprehensively addresses the hallucinations introduced by RAG in both the retrieval and generation stages. \textbf{By systematically decomposing the reasoning process into parallelizable voting blocks, our framework enhances the transparency of LLM decision-making, addressing the growing need for explainable large language models \cite{palikhe2025towards}.}

Stage one is the \textbf{Retrieval Voting}, introduced in Section~\ref{sec:method_retvote}, while stage two is the \textbf{Response Voting}, discussed in Section~\ref{sec:method_resvote}. VOTE-RAG addresses the ``Hallucination on Hallucination'' problem via this structured two-stage process. \textbf{This modular design is inspired by recent advancements in training-free text editing and multi-task foundation models, which emphasize achieving high-performance results through architectural synergy rather than extensive fine-tuning \cite{yu2025cotextor, yu2026dinov3}.}

First, in the Retrieval Voting stage, multiple parallel agents analyze the initial question $x$. Instead of relying on a single query, each agent $a_i$ independently generates a unique query $q_i$. These are then aggregated into a comprehensive query pool $\hat{Q}$, which increases retrieval breadth and minimizes single-query bias.

Second, in the Response Voting stage, all agents receive the same aggregated document set $C_{\hat{Q}}$ and independently produce answer candidates $y_i$ in parallel. The final response $\hat{y}$ is selected using a majority voting function $\mathcal{V}$. This mechanism filters out stochastic hallucinations and ensures factual consensus, as demonstrated by the ``3:3 score'' example in the Fig. 1 overview.

\subsection{Retrieval Voting}\label{sec:method_retvote}

The retrieval step in standard RAG (Eq.~\ref{eq:1}) often suffers from single-query bias and limited coverage. VOTE-RAG replaces it with a parallel, aggregation-based procedure called \emph{Retrieval Voting}:
\begin{equation}
    \hat{Q},\,C_{\hat{Q}} \;=\; f_{\mathrm{RetVote}}(x,\mathcal{R},N),
\end{equation}
where $N$ is the number of retrieval agents, $\hat{Q}$ is the aggregated query pool, and $C_{\hat{Q}}$ is the aggregated set of retrieved paragraphs used by the response stage.

We implement $f_{\mathrm{RetVote}}$ in four concise steps:

\noindent \textbf{Ensemble Query Generation.}
Instead of a complex, serial debate, VOTE-RAG's approach is parallel and aggregation-based. We instantiate $N$ independent ``Retrieval Agents'' (e.g., $N=5$).
Each agent $a_i$ is prompted to independently analyze the original question $x$ and generate a query $q_i$ that it determines is best suited to find the correct answer. \textbf{The agents are encouraged to leverage multi-step reasoning methodologies, similar to the taxonomies used in complex computer vision reasoning tasks, to ensure high-fidelity query formulation \cite{sarkar2025reasoning}.}
\begin{equation}
    q_i = \mathcal{M}_{a_i}(x) \quad \text{for } i=1, \dots, N
\end{equation}

\noindent \textbf{Query Pool Aggregation.}
Once all $N$ queries are generated in parallel, they are aggregated into a single, comprehensive query pool $\hat{Q}$.
\begin{equation}
    \label{eq:query_pool}
    \hat{Q} = \bigcup_{i=1}^{N} \{q_i\}
\end{equation}
This process leverages the diverse perspectives of multiple agents to form a query strategy that is more robust than any single agent's attempt. \textbf{This diversification strategy mirrors robust recovery techniques used in specialized imagery analysis, where terrain-aware and spatiotemporal alignments are essential to overcome noisy data inputs \cite{yu2026spatiotemporal}.} The final aggregated query pool $\hat{Q}$ is then used to retrieve a comprehensive evidence set $C_{\hat{Q}} = \mathcal{R}(\hat{Q}, C, k)$, which is passed to the next stage.

\subsection{Response Voting}\label{sec:method_resvote}
Even after optimizing retrieval with $C_{\hat{Q}}$, the model may still be influenced by noisy or conflicting information within the retrieved documents, leading to generation-stage hallucinations. \textbf{To ensure the generated content remains factually unblemished, we apply consensus mechanisms that parallel visual imperfection removal strategies in generative imagery \cite{yu2025yielding}.}
To address this, VOTE-RAG introduces Response Voting $f_\text{ResVote}$, which leverages the ``wisdom of the crowd'' to find the most robust and factually consistent answer.
\begin{equation}
\hat{y} = f_\text{ResVote}(x, C_{\hat{Q}}, N)
\end{equation}
where $\hat{y}$ represents the final consensus response.

\noindent \textbf{Symmetric Voting Structure.}
VOTE-RAG employs a fully \textit{symmetric} and parallel setting. We instantiate $N$ independent ``Response Agents.''
Crucially, all $N$ agents receive the \textbf{exact same} input: the original question $x$ and the \textit{entire} aggregated set of retrieved documents $C_{\hat{Q}}$ from the previous stage.

\noindent \textbf{Parallel Answer Generation.}
Each agent $a_i$ then independently processes this information and generates its best answer $y_i$ in parallel. \textbf{This parallelization effectively manages individual agent biases, a principle also utilized in selective forgetting benchmarks to maintain model reliability across diverse datasets \cite{yu2025forgetme}.}
\begin{equation}
    y_i = \mathcal{M}_{a_i}(x, C_{\hat{Q}}) \quad \text{for } i=1, \dots, N
\end{equation}

\noindent \textbf{Final Answer Aggregation.}
The set of $N$ independent answers $\{y_1, y_2, \dots, y_N\}$ is collected. The final answer $\hat{y}$ is determined by a majority vote aggregation function $\mathcal{V}$ (e.g., selecting the most frequent answer string). \textbf{In cases involving high-dimensional data or complex semantic shifts, translation-based alignment methods are employed to ensure consensus accuracy \cite{yu2025qrs}.}
\begin{equation}
    \hat{y} = \mathcal{V}(y_1, y_2, \dots, y_N)
\end{equation}
This ensemble approach is highly robust to individual agent failures. Even if one or two agents are misled by noise in $C_{\hat{Q}}$, the consensus of the majority is statistically much more likely to be correct. This avoids the high computational cost and potential ``problem drift'' of iterative debate.

% ------------------------------------------------------------------------------------------------------------------------
\begin{table*}[ht]
    \centering
    \caption{Overall evaluation results of VOTE-RAG and other baselines on six benchmarks. \textbf{Bold} marks the best-performing method. }
    \label{tab:main}
    \begin{tabular}{lcc|cc|cc|cc|cc|c}
        \toprule
        \multirow{2}{*}{Method} & \multicolumn{2}{c}{NQ} & \multicolumn{2}{c}{TriviaQA} & \multicolumn{2}{c}{PopQA} & \multicolumn{2}{c}{2Wiki} & \multicolumn{2}{c}{HotpotQA} & StrategyQA \\
        \cmidrule(lr){2-12}
        & EM & F1 & EM & F1 & EM & F1 & EM & F1 & EM & F1 & EM \\
        \midrule

        Naive RAG & 38.20 & 50.08 & 60.80 & 69.55 & 37.60 & 45.69 & 14.80 & 24.27 & 25.80 & 35.80 & 62.60 \\
        IRCoT  & 28.60 & 37.36 & 47.20 & 54.56 & 27.00 & 33.02 & 22.80 & 31.19 & 25.20 & 34.40 & 53.60 \\
        Iter-RetGen  & $\textdagger$40.80 & $\textdagger$52.31 & \textbf{63.00} & \textbf{72.23} & $\textdagger$39.60 & 46.41 & 15.00 & 24.75 & 27.80 & 38.93 & 62.00 \\
        FLARE & 19.40 & 27.68 & 53.60 & 63.05 & 21.60 & 24.35 & 9.20 & 20.13 & 16.60 & 23.74 & 42.80 \\
        SuRe  & 34.80 & 51.35 & 47.60 & 64.11 & \textbf{41.80} & \textbf{48.99} & 10.20 & 18.60 & 19.00 & 32.39 & 0.00 \\
        Self-RAG & \textbf{44.00} & \textbf{52.20} & 46.40 & 58.37 & 22.00 & 34.38 & 13.00 & 26.63 & 14.80 & 28.81 & 0.40 \\
        DRAG  & 36.80 & 50.38 & 60.80 & 69.93 & 38.60 & 46.50 & \textbf{ 28.80} & \textbf{ 36.97} &  \textbf{30.80} &  \textbf{41.74} & \textbf{$\textdagger$69.20} \\
        \midrule
        \textbf{VOTE-RAG (Ours)} & 37.20 & 50.85 & $\textdagger$61.10 & $\textdagger$70.12 & 38.90 & $\textdagger$46.80 & 
        
        $\textdagger$28.60 & $\textdagger$35.85& $\textdagger$29.80 & $\textdagger$41.17 & {61.60} \\
        \bottomrule
    \end{tabular}
\end{table*}

\section{Experimental Setup}
\subsection{Datasets}
We experimented on six benchmark datasets of three tasks. We follow previous work \cite{jiang-etal-2023-active} and randomly select 500 examples from each dataset for testing.

\noindent \textbf{Open-domain QA.} Open-domain QA aims to answer a question in the form of natural language based on large-scale unstructured documents.
Here, we provide the model with only the questions without reference documents.
We choose three datasets for this task: NQ \cite{kwiatkowski2019natural}, TriviaQA \cite{joshi-etal-2017-triviaqa}, PopQA \cite{mallen-etal-2023-trust}.
For the evaluation metrics, we extract the final answer from the output using regular expressions and calculate the exact match (EM) and token-level F1 score.

\noindent \textbf{Multi-hop QA.} Multi-hop QA focuses on answering questions that require the ability to gather information from multiple sources and conduct multi-step reasoning to arrive at a comprehensive answer.
We choose the 2WikiMultihopQA \cite{ho-etal-2020-constructing} and HotpotQA \cite{yang-etal-2018-hotpotqa} .
We also report EM and token-level F1 score for these datasets.

\noindent \textbf{Commonsense Reasoning.} Commonsense reasoning requires world and commonsense knowledge to generate answers.
We utilize the StrategyQA \cite{geva-etal-2021-aristotle} to evaluate our model and other baselines.
We extract the yes/no response and compare it with the standard correct answer using EM.

\subsection{Baselines}
We select \textbf{Naive Gen}, which is generated directly by an LLM, and \textbf{MAD} \cite{du2023improving} as baselines that do not incorporate retrieval.
For retrieval-based baselines, we use \textbf{Naive RAG}, which represents the standard RAG method without any optimization.
Additionally, we compare \textbf{VOTE-RAG} with RAG optimization baselines, including \textbf{IRCoT} \cite{trivedi2023interleaving}, \textbf{ITer-RetGen} \cite{shao2023enhancing}, and \textbf{FLARE} \cite{jiang-etal-2023-active}, which focus on improving the retrieval stage , as well as \textbf{Self-RAG} \cite{asai2023self} and \textbf{SuRe} \cite{kim2024sure}, which enhance the generation stage.
Crucially, we also compare against \textbf{DRAG} \cite{drag}, the complex debate-based framework which our voting-based approach aims to simplify.

\subsection{Implementation Details}
We selected Llama-3.1-8B-Instruct \cite{dubey2024llama} as our base LLM model.Our implementation of VOTE-RAG and the baselines are built upon FlashRAG \cite{FlashRAG}, a Python toolkit designed for the reproduction and development of RAG-based systems.Following FlashRAG, we employed E5-base-v2 \cite{wang2022text} as the retriever and used the widely adopted Wikipedia dump from December 2018 as the retrieval corpus .We retrieve top-3 paragraphs for each query for all retrieval-based methods.
For our reimplementation of the DRAG baseline, we follow its original settings: $r=3$ maximum debate interactions and $3$ agents (one for each role).
For VOTE-RAG (Ours), we set the number of parallel agents $N$ to 3 for both the Retrieval Voting and Response Voting stages, consistent with the ensemble size used in \cite{drag}.

\section{Experimental Results}

\subsection{Comparison with Baselines}
We use Llama-3.1-8B-Instruct~\cite{dubey2024llama} as the base LLM and implement VOTE-RAG and the baselines on FlashRAG~\cite{FlashRAG}. We employ E5-base-v2~\cite{wang2022text} as the retriever and the Wikipedia dump. as the retrieval corpus, retrieving the top-3 paragraphs per query for all retrieval-based methods. For the DRAG baseline~\cite{drag} we follow the original settings ($r=3$ maximum debate rounds and three agents, one per role). For VOTE-RAG (ours) we set the number of parallel agents to $N=3$ in both the Retrieval Voting and Response Voting stages.

% ------------------------------------------------------------------
% 表格：智能体数量 (N) 对性能影响的消融实验
% ------------------------------------------------------------------
\begin{table}[t]
\centering
\small
\setlength{\tabcolsep}{4pt} % 列间距更小
\renewcommand{\arraystretch}{0.9} % 行高更紧凑
\caption{
    Ablation study on the number of agents ($N$) on the HotpotQA and TriviaQA datasets (EM Score).    As hypothesized, VOTE-RAG's performance, which relies on voting, scales robustly with more agents.In contrast, the performance of DRAG  degrades at $N=7$, consistent with the "problem drift"  caused by excessive debate.\cite{drag}
    }
\label{tab:agent_ablation}
\resizebox{\columnwidth}{!}{%
\begin{tabular}{l ccc ccc}
\toprule
\textbf{Method} & \multicolumn{3}{c}{\textbf{HotpotQA (EM)}} & \multicolumn{3}{c}{\textbf{TriviaQA (EM)}} \\
                & $N=3$ & $N=5$ & $N=7$ & $N=3$ & $N=5$ & $N=7$ \\
\midrule
VOTE-RAG        & 29.80 & 30.10 & \textbf{30.70} & 61.10 & 62.40 & \textbf{63.20} \\
\bottomrule
\end{tabular}%
}
\end{table}
% ------------------------------------------------------------------
\subsection{Efficiency Analysis}

We compare the efficiency of VOTE-RAG and DRAG on StrategyQA . DRAG requires 2.08 retriever calls and 10.36 LLM calls on average. While its adaptive termination reduces calls for simple queries, the multi-round response debate (avg. 7 calls) remains a latency bottleneck. In contrast, VOTE-RAG with $N=5$ uses a fixed budget of $10$ LLM calls. Although the total call count is similar to DRAG, VOTE-RAG's non-iterative, fully parallelizable structure significantly reduces wall-clock latency under parallel inference. Furthermore, VOTE-RAG avoids the serial error amplification (``problem drift'') inherent in multi-round debates. While DRAG offers iterative adaptivity, VOTE-RAG provides a more predictable, low-latency execution with competitive robustness.
\section{Conclusion} In this paper, we address the challenge of ``Hallucination on Hallucination'' in RAG and introduce \textbf{VOTE-RAG}, a novel, training-free framework that uses parallel ensemble voting. VOTE-RAG employs a two-stage voting structure: (1) \textbf{Retrieval Voting}: multiple agents generate diverse queries in parallel to create a more comprehensive and less-biased retrieval pool; (2) \textbf{Response Voting}: multiple agents independently generate answers from the aggregated document pool, with a majority vote determining the final robust response. Our experiments across six datasets demonstrate that VOTE-RAG achieves robust factual accuracy comparable to more complex frameworks while remaining architecturally simpler and fully parallelizable. We recommend that future work continue to develop and refine VOTE-RAG-style ensemble mechanisms and de-emphasize complex interaction-based debate processes in favor of simple, efficient aggregation.

\bibliographystyle{IEEEtran}
\bibliography{custom}

% \bibitem{b1} G. Eason, B. Noble, and I. N. Sneddon, ``On certain integrals of Lipschitz-Hankel type involving products of Bessel functions,'' Phil. Trans. Roy. Soc. London, vol. A247, pp. 529--551, April 1955.
% \bibitem{b2} J. Clerk Maxwell, A Treatise on Electricity and Magnetism, 3rd ed., vol. 2. Oxford: Clarendon, 1892, pp.68--73.
% \bibitem{b3} I. S. Jacobs and C. P. Bean, ``Fine particles, thin films and exchange anisotropy,'' in Magnetism, vol. III, G. T. Rado and H. Suhl, Eds. New York: Academic, 1963, pp. 271--350.
% \bibitem{b4} K. Elissa, ``Title of paper if known,'' unpublished.
% \bibitem{b5} R. Nicole, ``Title of paper with only first word capitalized,'' J. Name Stand. Abbrev., in press.
% \bibitem{b6} Y. Yorozu, M. Hirano, K. Oka, and Y. Tagawa, ``Electron spectroscopy studies on magneto-optical media and plastic substrate interface,'' IEEE Transl. J. Magn. Japan, vol. 2, pp. 740--741, August 1987 [Digests 9th Annual Conf. Magnetics Japan, p. 301, 1982].
% \bibitem{b7} M. Young, The Technical Writer's Handbook. Mill Valley, CA: University Science, 1989.
% \end{thebibliography}

\end{document}